\documentclass[a4paper]{article}

\usepackage{INTERSPEECH2018}
\usepackage{hyperref}
\usepackage{makecell}

\title{Attentive Sequence-to-Sequence Learning for \\
  Diacritic Restoration of Yor{\`u}b{\'a} Language Text}
\name{Iroro Fred \d{\`O}n\d{\`o}m\d{\`e} Orife}
\address{Niger-Volta Language Technologies Institute}
\email{iroro@alumni.cmu.edu}

\begin{document}

\maketitle
\begin{abstract}
Yor{\`u}b{\'a} is a widely spoken West African language with a writing system rich in tonal and orthographic diacritics. With very few exceptions, diacritics are omitted from electronic texts, due to limited device and application support. Diacritics provide morphological information, are crucial for lexical disambiguation, pronunciation and are vital for any Yor{\`u}b{\'a} text-to-speech (TTS), automatic speech recognition (ASR) and natural language processing (NLP) tasks. Reframing Automatic Diacritic Restoration (ADR) as a machine translation task, we experiment with two different attentive Sequence-to-Sequence neural models to process undiacritized text. On our evaluation dataset, this approach produces diacritization error rates of less than 5\%. We have released pre-trained models, datasets and source-code as an open-source project to advance efforts on Yor{\`u}b{\'a} language technology.
\end{abstract}
\section{Introduction}

Yor{\`u}b{\'a} is a tonal language spoken by more than 40 Million people in the countries of Nigeria, Benin and Togo in West Africa. There are an additional million speakers in the African diaspora, making it the most broadly spoken African language outside Africa \cite{yoruba_language}. The phonology is comprised of eighteen consonants \emph{({b}, {d}, {f}, {g}, {gb}, {h}, {j}, {k}, {l}, {m}, {n}, {p}, {r}, {s}, \d{s}, {t}, {w}, y)}, seven oral vowel \emph{({a}, {e}, \d{e}, \d{i}, {o}, \d{o}, {u})} and five nasal vowel phonemes \emph{({an}, \d{e}{n}, {in}, \d{o}{n}, {un})} with three kinds of tones realized on all vowels and syllabic nasal consonants \emph{({\'m}, {\'n})} \cite{akinlabi2004sound}. Accordingly, Yor{\`u}b{\'a} orthography makes significant use of tonal diacritics to signify tonal patterns, and orthographic diacritics like underdots for various language sounds \cite{adegbola2012quantifying}. For example, \emph{\textbf{\d{e}}} signifies a half-open vowel, while \emph{\textbf{\d{s}}} represents a palatoalveolar fricative \cite{wells2000orthographic}.

On modern computing platforms, the vast majority of Yor{\`u}b{\'a} text is written in plain ASCII, without diacritics. This presents grave problems for usage of the standard orthography via electronic media, which has implications for the unambiguous pronunciation of Yor{\`u}b{\'a}'s lexical and grammatical tones by both human speakers and TTS systems. Improper handling of diacritics also degrades the performance of document retrieval via search engines and frustrates every kind of Natural Language Processing (NLP) task, notably machine translation to and from Yor{\`u}b{\'a} \cite{asubiaro2014effects}. Finally, correct diacritics are mandatory in reference transcripts for any Automatic Speech Recognition (ASR) task.

\subsection{Ambiguity in non-diacritized text}
Automatic Diacritic Restoration (ADR), which goes by other names such as Unicodification \cite{scannell2011statistical} or deASCIIfication \cite{arslan2016deasciification} is a process which attempts to resolve the ambiguity present in undiacritized text. Table~\ref{tab:examples} shows diacritized forms for each non-diacritic character. 

 Undiacritized Yor{\`u}b{\'a} text has a high degree of ambiguity \cite{adegbola2012quantifying, asahiah2017restoring, de2007automatic}. Adegbola et al. state that for ADR the ``prevailing error factor is the number of valid alternative arrangements of the diacritical marks that can be applied to the vowels and syllabic nasals within the words" \cite{adegbola2012quantifying}. For our training corpus of 1M words, we quantify the ambiguity by the percentage of all words that have diacritics, 85\%; the percentage of unique non-diacritized word types that have two or more diacritized forms, 32\%, and the lexical diffusion or \emph{LexDif} metric, which conveys the average number of alternatives for each non-diacritized word, 1.47. 
  \begin{table}[h]
  \caption{Characters with their non-diacritic forms }
  \label{tab:examples}
  \centering
  \begin{tabular}{lcl}
    \toprule
    \multicolumn{2}{c}{\textbf{Characters}} & \textbf{Examples}  \\
    \midrule
    {\`a} {\'a} \v{a} & \textbf{a} & gb{\`a} \emph{(spread)}, gba \emph{(accept)}, gb{\'a} \emph{(hit)}    \\  
    {\`e} {\'e} \d{e} \d{\`e} \d{\'e} & \textbf{e} & es{\'e} \emph{(cat)}, {\`e}s{\`e} \emph{(dye)}, \d{e}s\d{\`e} \emph{(foot)} \\
    {\`i} {\'i} & \textbf{i} & {\`i}l{\'u} \emph{(town)}, ilu \emph{(opener)}, {\`i}l{\`u} \emph{(drum)}\\  
    {\`o} {\'o} \d{o} \d{\`o} \d{\'o} \v{o} & \textbf{o} & \d{o}k\d{\'o} \emph{(hoe)}, \d{\`o}k\d{\`o} \emph{(spear)}, \d{o}k\d{\`o} \emph{(vehicle)}\\  
    {\`u} {\'u} \v{u} & \textbf{u} & mu \emph{(drink)}, m{\`u} \emph{(sink)},  m{\'u} \emph{(sharp)} \\
    \midrule
    {\`n} {\'n} \={n} & \textbf{n} & {n} \emph{(I)}, {\'n} (continuous aspect marker) \\  
    \d{s} & \textbf{s} &  {s}{\'a} \emph{(run)}, \d{s}{\'a} \emph{(fade)}, \d{s}{\`a} \emph{(choose)} \\  
    \bottomrule
  \end{tabular}
\end{table}
Further, 64\% of all unique, non-diacritized monosyllabic words possess multiple diacritized forms \cite{oluseye2003yoruba, delano1969dictionary}. When we consider the distribution of ambiguity over grammatical function, we recognize the added difficulty of tasks like the lexical disambiguation of non-diacritized Yor{\`u}b{\'a} verbs, which are predominantly monosyllabic.

\begin{table}[h]
\caption{Tonal Changes}
\label{tab:tonal_changes}
\centering
\begin{tabular}{clll}
   \toprule
   \textbf{Verb}  & \textbf{Phrase} & \textbf{Translation}  & \textbf{Tone}\\
   \midrule
   t{\`a}  & o t{\`a} a & he sells it & Low \\ 
                    & o ta i\d{s}u, o ta\d{s}u & he sells yams & Mid\\  
   \midrule
    t{\`a} & o ta {\`i}w{\'e} & he sells books & Mid \\ 
                    & o t{\`a}w{\'e} & he sells books & Low\\  
   \bottomrule
 \end{tabular}
\end{table}

Finally, there are tonal changes, which are rules about how tonal diacritics on a specific word are \emph{altered} based on context. In the first example in Table \ref{tab:tonal_changes}, a low-tone verb \textbf{t{\`a}} \emph{(to sell)} becomes mid when followed or combined with a noun object \textbf{i\d{s}u}. In the second example, the same low-tone verb retains its tone if there is an elision with a noun object \textbf{{\`i}w{\'e}}, otherwise the verb becomes a mid-tone \cite{delano1969dictionary}.

\subsection{Seq2Seq Approach} 
To date, methods to solving Yor{\`u}b{\'a} ADR have used memory-based or Na{\"i}ve Bayes classifers on n-gram features. Efforts have focused on the word-level or mixed-models, rather than purely character-level models based on results that indicate ``tonal diacritics can simply not be solved on the level of the grapheme" \cite{de2007automatic}. With some studies using corpora as small as 5k words from a 3.5k lexicon \cite{scannell2011statistical}, the dearth of accurate diacritized electronic text has been the object of study as well as a limiting factor on progress \cite{adegbola2012quantifying}. 

Recently, neural machine translation (NMT) has emerged as the state-of-the-art approach to solving automatic inter-language machine translation. Expressing ADR as a machine translation problem, we treat undiacritized text and diacritized text as source and target languages respectively in a NMT formulation. Our contributions are as follows:
\begin{itemize}
\item We propose two different NMT approaches, using soft-attention and self-attention sequence-to-sequence (seq2seq) models \cite{bahdanau2014neural, vaswani2017attention}, to rectify undiacritized Yor{\`u}b{\'a} text.
\item We release the training datasets, pre-trained models, source code and reproducible results into the public domain as an open-source project.
\end{itemize}
This paper is organized as follows. Section 2 reviews previous work in Yor{\`u}b{\'a} ADR. In section 3, we review algorithms for seq2seq learning. In section 4 and 5, we present our experimental setup and results. In section 6, we conclude with a review of applications and future directions.

\section{Related work}

ADR is an active field of study with on-going efforts in a wide variety of languages, including Czech, Polish, Romanian and Hungarian \cite{novak2015automatic, tufics2008diac+, mihalcea2002letter}, Turkish \cite{arslan2016deasciification}, Arabic \cite{khorsheed2012hmm, belinkov2015arabic, nelken2005arabic, schlippe2008diacritization}, M{\=a}ori \cite{cocks2011word}, Uyghur \cite{tursun2017noisy}, Urdu \cite{raza2010automatic}, Vietnamese \cite{pham2017use, luu2012pointwise, do2013machine}, as well as West African languages like Igbo \cite{ezeani2016automatic} and Yor{\`u}b{\'a} \cite{adegbola2012quantifying, scannell2011statistical, asahiah2017restoring, de2007automatic}. 

ADR investigations have relied on ruled-based morphological analyzers \cite{novak2015automatic, raza2010automatic}, or used a variety of statistical learning techniques including conditional random fields (CRFs) \cite{schlippe2008diacritization}, support vector machines (SVMs) \cite{luu2012pointwise}, Hidden Markov Models (HMMs) \cite{khorsheed2012hmm}, finite state transducers \cite{nelken2005arabic}, n-gram models \cite{raza2010automatic} and recurrent neural networks \cite{belinkov2015arabic}. Parallel corpora, statistical machine translation (SMT) techniques have also been used \cite{novak2015automatic, schlippe2008diacritization}. We will focus our review specifically on Yor{\`u}b{\'a} and ADR techniques using NMT sequence-to-sequence models.

\subsection{Yor{\`u}b{\'a} ADR}

The complexity of the Yor{\`u}b{\'a} ADR task is comparable to languages like Vietnamese where over 90\% of words contain diacritics, of which some 80\% are ambiguous without diacritics \cite{pham2017use, do2013machine}. In a contrastive study of seven resource-scarce African languages and better resourced European and Asian languages, DePauw et al. \cite{de2007automatic} report \emph{LexDif} scores of 1.26 and 1.21 respectively for Yor{\`u}b{\'a} and Vietnamese, citing the similar disambiguation tasks due to the marking of phonemic variants and tonal characteristics. In contrast, French and Romanian corpora, where approximately a fifth and two-fifths of words contain diacritics, received \emph{LexDif} scores of 1.04 and 1.05 respectively \cite{simard1998automatic}. 

For Yor{\`u}b{\'a},  DePauw et al. trained a Tilburg Memory-Based Learner (TiMBL) classifier that predicted the correct diacritic based on the local graphemic context. Memory-based learning is a variant of the classical k-Nearest Neighbors (\emph{k}-NN) approach to classification, common for NLP tasks. Trained on a corpus of 65.6k words, their best word-level accuracy was 76.8\%. 

Scannell \cite{scannell2011statistical} implemented a Na{\"i}ve Bayes classifier using word and character-level models. For character-level prediction, each ambiguous character was treated as a separate classification problem, disregarding any previous diacritization. For word-level, two lexicon lookup methods were used: The first replaces ambiguous words with the most frequent candidate, while the second uses a bigram model to determine the output. The corpus was a meager 5k words from a 3.5k word lexicon. Word-level models significantly outperformed character-level models with a top word-level accuracy of 75.2\%.

Investigating the effect of corpus size on ADR accuracy, Adegbola et al. \cite{adegbola2012quantifying} used a Na{\"i}ve Bayes classifier based on word trigram probabilities and linear interpolation for smoothing. Trained on 100k words with a 7.5k lexicon, the best word-level ADR result was 70.5\%.

Asahiah et al. \cite{asahiah2017restoring} tackle a subset of the full ADR task, uniquely focusing on tone mark restoration, not orthographic diacritics. On a corpus of 250k words, using a TiMBL classifier and syllables as the unit of restoration, they report mean accuracies at the syllable and word-level of 98\% and 93\% respectively.
\\
\\
Note that in each of the four studies, the evaluation corpora are private and the results from each are not directly comparable to the others.

\subsection{Sequence-to-sequence learning for ADR}

There have been only two investigations, to our knowledge, that use sequence-to-sequence learning for ADR tasks. Pham et al. \cite{pham2017use} compared the performance of a standard RNN Encoder-Decoder pair on a Vietnamese ADR task. The results on a dataset of 180k sentence pairs yielded accuracy scores of 96.15\%, in comparison to 97.32\% for the state-of-the-art phrase-based approach for Vietnamese.

For Uyghur language text normalization, Tursun et al. \cite{tursun2017noisy} use a non-attentive sequence-to-sequence model for character-level restoration. On a small corpus of 1372 words (226 sentences), they report accuracy of 65.7\% for the sequence-to-sequence model and in comparison, 64.9\%, for a noisy channel model.

\section{Sequence-to-sequence learning}

Neural machine translation (NMT) is a recent approach to machine translation that uses models belonging to a family of encoder-decoders neural networks, trained to maximize the probability of a correct translation given a source sentence \cite{sutskever2014sequence, cho2014learning}. Recurrent neural networks (RNNs) are a common choice for encoders and decoders because they can learn a probability distribution over a sequence of symbols by being trained to predict the next symbol in the sequence \cite{cho2014learning}. 

In the basic \emph{RNN Encoder-Decoder} architecture, an RNN-Encoder operates on a variable-length source sequence $\mathbf x$ $=(x_1,...,x_{T_x})$ to generate a fixed-length summary context vector $\mathbf c$ from its sequence of hidden states. At time step $t$, the encoder's hidden state $h^e_t \in {\Bbb R^n}$ is updated by
\begin{equation}
  h^e_t = f_{enc}\ (x_t, \mathbf h^e_{t-1})
  \label{eq1}
\end{equation}
and
\begin{equation}
  \mathbf c = g_{enc}\ (\{\mathbf h^e_1,...,\mathbf h^e_{T_x} \})
  \label{eq2}
\end{equation}
 $f_{enc}$ and $g_{enc}$ are recurrent non-linear activation functions, as simple as a logistic sigmoid or as complex as a gated recurrent unit (GRU) or long short-term memory (LSTM) unit \cite{bahdanau2014neural}. The RNN-Decoder is another RNN trained to generate an output sequence $\mathbf y$ $=(y_1,...,y_{T_y})$, by predicting the next word $y_{t}$ from previously predicted words, its hidden state $\mathbf h^d$ and the context vector $\mathbf c$. At time step $t$, the decoder's hidden state $\mathbf h^d_t$ is updated by
\begin{equation}
  \mathbf h^d_t = f_{dec}\ (y_{t-1}, \mathbf h^d_t, \mathbf c)
  \label{eq3}
\end{equation}
and the conditional distribution of the next symbol is
\begin{equation}
  p(y_t|\{y_1,...,y_{t-1}\}, \mathbf x) = softmax(g_{dec}(\mathbf h^d_t))
  \label{eq4}
\end{equation}
where $f_{dec}$ is another recurrent non-linear activation function like an LSTM. $g_{dec}$ is a transformation function that returns a vocabulary-sized vector, from which the softmax function outputs the probability of $y_t$ \cite{luong2015effective}. The joint training objective is to minimize the conditional negative log-likelihood
\begin{equation}
 	J = - \frac{1}{N}\sum\limits_{n=1}^{N}\sum\limits_{t=1}^{T^{(n)}_y} log\ p(y_t = y^{(n)}_t|y^{(n)}_{<t}, \mathbf x^{(n)})
  \label{eq5}
\end{equation}
where $N$ is the number of parallel training sentence pairs, and $\mathbf x^{(n)}$ and $y^{(n)}_{t}$ are the source sentence and $t$th target symbol in the $n$th pair respectively \cite{luong2015effective}. Though initial RNN-Decoders only observed the last encoder hidden state, with context vector $\mathbf c = h^e_{T_x}$, variants of this architecture differ in the type of RNN and how the context vector $\mathbf c$ is derived \cite{bahdanau2014neural, cho2014learning}. 

\subsection{Soft-attention}

The first architecture used in this study is based on the work of Bahdanau et al. \cite{bahdanau2014neural}, which extends the RNN-Encoder-Decoder design with an attention mechanism that allows the decoder to observe \emph{different} source words for each target word. A bi-directional RNN is used for the encoder network to create representations that consider both past and future inputs. The state $\mathbf h^e_j$ corresponds to the concatenation of states produced by the forward and backward RNNs: $\mathbf h^e_j = [\overrightarrow{h^e_j} ; \overleftarrow{h^e_j}]$. At each decoding time step, the context vector $\mathbf c_{i}$ is now computed as a weighted sum of the encoder hidden states
\begin{equation}
 	c_{i} = \sum\limits_{j=1}^{T_y} {\alpha}_{{i}{j}}{\mathbf h^e_j}
  \label{eq6}
\end{equation}
where ${\alpha}_{{i}{j}} = softmax(e_{{i}{j}})$ and $e_{{i}{j}} = att(\mathbf h^d_{i-1}, \mathbf h^e_j)$. The magnitude of the attentional weight ${\alpha}_{{i}{j}}$ represents how important the $j$th source token $x_j$ is to the $i$th target token $y_i$. $e_{{i}{j}}$ is an unnormalized compatability score between the encoder state $\mathbf h^e_j$ (inputs around position $j$) and the decoder state $\mathbf h^d_{i-1}$ (outputs around position $i$) \cite{bahdanau2014neural, britz2017massive}. This score can be computed most simply as the dot-product between the vectors, where the matrices $W$ transform the encoder and decoder states into a same-sized representation. Bahdanau et al. define $att$ as a small feed-forward neural network with a single hidden layer with $v_a$ as an additional weight matrix \cite{bahdanau2014neural}.
\begin{equation}
 	 att(\mathbf h^d_{i-1}, \mathbf h^e_j) = 
 	\begin{cases}
   	\langle W_d \mathbf h^d_{i-1}, W_e \mathbf h^e_j \rangle  & \mathbf {dot} \\
	v_a^\top tanh(W_d \mathbf h^d_{i-1} + W_e \mathbf h^e_j)  & \mathbf {add}
    \end{cases}	
   \label{eq7}
\end{equation}
The $att$ network is jointly trained end-to-end with the other components of the system by minimizing the conditional negative log-likelihood of the target words. The conditional distribution of the next decoded symbol, as expressed in Equation \ref{eq4}, remains the same. The attention mechanism only affects the computation of the context vector $c_{i}$, effectively relieving the encoder from the trouble of encoding all source information into a fixed-length vector. Rather, distributed information throughout the encoder hidden states can be selectively retrieved by the decoder, yielding both performance and scalability benefits.
\begin{figure}[h]
  \centering
  \includegraphics[width=3.3in]{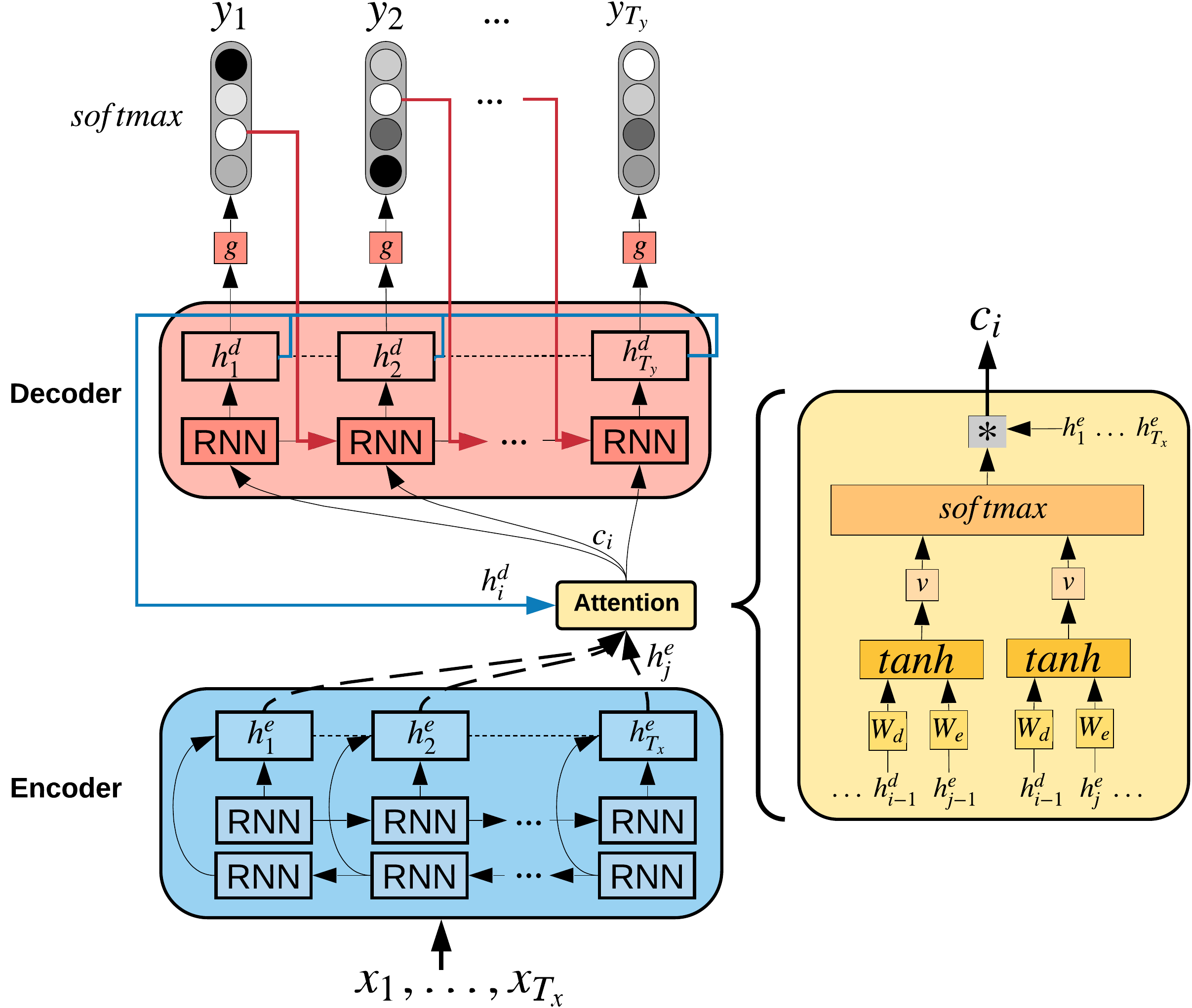}
  \caption{Soft-attention}
  \label{fig:Bahdanau_soft_attention}
\end{figure}

\subsection{Self-attention}

The second attentive seq2seq architecture employed in this study aims to improve on limitations of RNNs, i.e. high computational complexity and non-parallelizeable computation. For both the encoder and decoder, the Transformer model, proposed by Vaswani et al. \cite{vaswani2017attention}, employs stacks of self-attention layers in lieu of RNNs. Intuitively, self-attention, or intra-attention, computes a representation of a single sequence, modeling dependencies between words from the same sequence. It serves the same purpose as the bi-directional RNN, capturing more directly and efficiently, the relevant context for each word in a sequence. 

Self-attention layers use multiple attention heads, with each head mapping a source sequence $\mathbf x$ $=(x_1,...,x_{T_x})$ into a new sequence of the same length $\mathbf z$ $=(z_1,...,z_{T_x})$. The source $x$ is linearly transformed into queries, keys and values matrices using parameter matrices $W^Q$, $W^K$, $W^V$ for each layer and each attention head. An output element $z_i$ is computed as
\begin{equation}
 	z_{i} = \sum\limits_{j=1}^{T_x} {\alpha}_{{i}{j}}({x_j}W^V)
  \label{eq6}
\end{equation}
where as with soft-attention above, ${\alpha}_{{i}{j}} = softmax(e_{{i}{j}})$ and 
\begin{equation}
 	 e_{ij} = \frac{1}{\sqrt{d_z}}(x_i W^Q)(x_j W^K)^\top
   \label{eq8}
\end{equation}
The scaled-dot product function in Equation \ref{eq8} computes a compatibility score $e_{ij}$, between queries and keys. $h$ parallel heads are used to attend to different parts of the value vectors, concatenating the output of each head to form a single output vector. Because self-attention layers are inherently invariant to sequence ordering, explicit positional encodings are concatenated along with the input. Residual connections also help distribute position information to higher layers.

Overall, the Transformer encoder is composed of 6 identical layers, each combining a self-attention sub-layer with a fully-connected feed-forward network. The decoder has a similar construction, adding a third sub-layer, which does multi-head attention over the output of the encoder stack, i.e. encoder-decoder attention. The joint training objective remains the same as in the previous sequence-to-sequence architectures.

\section{Experimental setup}

We review our dataset collection and preprocessing methods then discuss tools and techniques for ADR model training.

\subsection{Text preparation}

To prepare source and target texts for parallel training, we obtained a very small but fully diacritized text from the Lagos-NWU conversational speech corpus by Niekerk, et. al \cite{niekerk2012tone}. We also created our own medium-sized corpus by web-crawling the two Yor{\`u}b{\'a}-language websites with full diacritics, a current events blog and an online Bible. 

 Collecting data from varied sources this way is resourceful but also introduces covariate shift between the data subsets. This necessitated text preprocessing of all web-crawled text to ensure common character sets with minimal punctuation. We also cleaned up texts to ensure consistent, error-free diacritization, splitting lines on full-stops to give one sentence per line. To ensure our splits are drawn from similar distributions, we combined all text, shuffled and split utterances into a ratio of 80\%, 10\%, 10\%, for training, test and dev sets respectively.   
 \begin{table}[h]
  \caption{Training data subsets}
  \label{tab:training_datasets}
  \centering
  \begin{tabular}{lll}
    \toprule
    \textbf{\# words} & \textbf{Source URL}  & \textbf{Description} \\
    \midrule
    24,868 & rma.nwu.ac.za  & Lagos-NWU corpus \\  
    50,202 & theyorubablog.com & language blog\\  
    910,401 & bible.com & online bible webite \\
    \bottomrule
  \end{tabular}
\end{table}

Next, all characters in the texts were dispossessed of their diacritics. This entailed converting diacritized text to Unicode Normalization Form Canonical Decomposition (NFD) which separates a base character from its diacritics. We then filtered out all the \emph{UnicodeCategory.NonSpacingMark} characters, which house the diacritic modifications to a character. This yielded two sets of text, one stripped of diacritics (the source) and the other with full diacritics (the training target). To better understand the dataset split, we computed a perplexity of 575.6 for the test targets with a language model trained over the training targets \cite{stolcke2002srilm}. The \{source, target\} vocabularies for training and test sets have \{11857, 18979\} and \{4042, 5641\} word types respectively.

\subsection{Training}
We built the soft-attention and self-attention models with the Python 3 implementation of OpenNMT, an open-source toolkit created by the Klein et al. \cite{opennmt}. Our choice of programming language included practical engineering considerations, including Unicode support and modern asynchronous execution. In a training framework, we needed an expressive, extensible interface, documentation and Tensorboard integration. Our training hardware configuration was a standard AWS EC2 p2.xlarge instance with a NVIDIA K80 GPU, 4 vCPUs and 61GB RAM. Training the various models took place over the course of a few days.

Experiments varied over the dimensionality of the word embedding layers, the number and size of the RNN layers, the attention type ($tanh$, $dot$) and optimizer's hyperparameters. Soft-attention training usually converged within 50 epochs, using the Adam optimizer with learning rate decay. The Transformer model however needed only 25 epochs. We selected the top 5 models with the best results on the held-out test set in Table \ref{tab:results}. 

\section{Results}
To evaluate the performance of our ADR models, we computed the accuracy score as the ratio of correct words restored to all words. We calculate the perplexity of each model's predictions based on the test set targets.
 \begin{table}[h]
  \caption{Training \& Test Accuracy and Perplexity}
  \label{tab:results}
  \centering
  \begin{tabular}{cccccc}
    \toprule
    \textbf{Attention} & \textbf{Size} & \textbf{RNN} & \textbf{Train\%} & \textbf{Test\%} &\textbf{PPL} \\
    \midrule
    soft + dot & 2L 512 & LSTM & 96.2 & 90.1 & 1.68 \\
    soft + add & 2L 512 & LSTM & 95.9 & 90.1 & 1.85 \\
	soft + tanh & 2L 512 & GRU & 96.2 & 89.7 & 1.83 \\ 
	soft + tanh & 1L 512 & GRU  & 97.8 & 89.7 & 1.86 \\ 
	\midrule
    self & 6L 512   & -  & 98.5 & 95.4 & 1.32 \\
    \bottomrule
  \end{tabular}
\end{table}
The type of RNN or attention did not make much difference in accuracy, with GRUs and $tanh$ attention being as accurate as dot-product attention with LSTMs. We suspect that attention type will have a greater influence on accuracy when training on a much larger corpus. 

Early in the training process, models exhibited word order errors in addition to  partial and incorrect diacritizations. As learning slowed, models from both attention architectures were able to satisfactorily learn long-range dependencies and the correct context for most tonal and orthographic diacritics, with the Transformer model being slightly more accurate. 

 \begin{table}[h]
  \caption{A sample diacritization from the test set}
  \label{tab:results_samples}
  \centering
  \begin{tabular}{rl}
    \toprule
    \textbf{source} & emi ni oye ju awon agba lo nitori mo gba eko re\\
    \textbf{target} & {\`e}mi ni {\`o}ye j{\`u} {\`a}w\d{o}n {\`a}gb{\`a} l\d{o} n{\'i}tor{\'i} mo gba \d{\`e}k\d{\'o} r\d{e} \\
    \textbf{prediction} & {\`e}mi ni {\`o}ye \underline{\textbf{j{\`u}} {\`a}w\d{o}n {\`a}gb{\`a} \textbf{l\d{o}}} n{\'i}tor{\'i} mo gba \d{\`e}k\d{\'o} r\d{e}\\
    \midrule
    \textbf{source} & emi yoo si si oju mi si juda \\
    \textbf{target} & {\`e}mi y{\'o}{\`o} s{\`i} s{\'i} oj{\'u} mi s{\'i} il{\'e} j{\'u}d{\`a} \\
    \textbf{prediction} & {\`e}mi y{\'o}{\`o} s{\`i} s{\'i} oj{\'u} mi s{\'i} il{\'e} j{\'u}d{\`a} \\
    \midrule
    \textbf{source} & bi iji ti n fe eefin lo ki o fe won lo \\
    \textbf{target} & b{\'i} {\`i}j{\`i} ti {\'n} f\d{\'e} {\`e}{\'e}f{\'i}n l\d{o} k{\'i} {\'o} f{\'e} w\d{o}n l\d{o} \\
    \textbf{prediction} & b{\'i} {\`i}j{\`i} ti {\'n} f\d{\'e} {\`e}{\'e}f{\'i}n l\d{o} k{\'i} {\'o} f{\'e} w\d{o}n l\d{o} \\
    \midrule
	\textbf{source} & oun ko si ni ko ile ini re sile \\
	\textbf{target} &   {\`o}un k{\`o} s{\`i} n{\'i} \textbf{k\d{o}} il\d{\`e} {\`i}n{\'i}  r\d{\`e} s{\'i}l\d{\`e}\\
	\textbf{prediction} &  {\`o}un k{\`o} s{\`i} n{\'i} \textbf{k\d{\'o}} il\d{\`e} {\`i}n{\'i}  r\d{\`e} s{\'i}l\d{\`e} \\ 
    \bottomrule
  \end{tabular}
\end{table}
For example, the verbs \textbf{b{\`a}j\d{\'e}} (to spoil), or \textbf{j{\`u}l\d{o}} (to be more than) are discontinuous morphemes, or splitting verbs. In Table \ref{tab:results_samples}, the first example shows the model has learnt the diacritics necessary for \textbf{l\d{o}} following a previously predicted \textbf{j{\`u}}. 

\begin{figure}[h]
  \centering
  \includegraphics[width=3.4in]{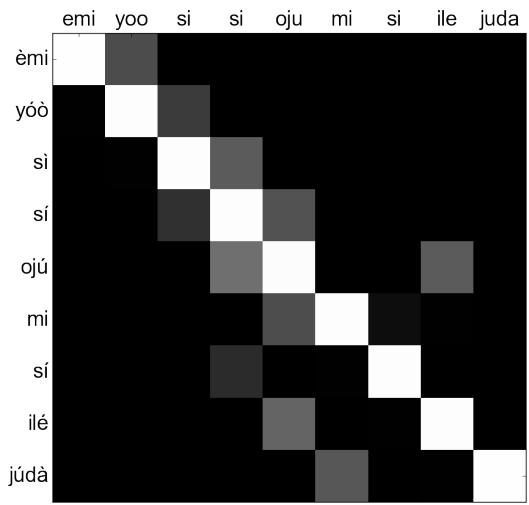}
  \caption{Attention weight matrix}
  \label{fig:attention_weights_emiyoo}
\end{figure}

In Figure \ref{fig:attention_weights_emiyoo} we plot attention weight matrices for the second example in Table \ref{tab:results_samples}.
As each predicted word considers its undiacritized form, the attention weights matrices feature a prominent diagonal. We note the ambiguity of the undiacritized \textbf{si} in this context, with two diacritized forms, \textbf{s{\`i}}, \textbf{s{\'i}}, we observe that the third instance \textbf{s{\'i}} attends to the previous \textbf{s{\'i}} and that the first two attend to each other. Finally, as this phrase roughly translates as \emph{I will keep my eyes on the house of Judah}, it is interesting to note that the nouns, \textbf{oj{\'u}} (eyes) and \textbf{il{\'e}} (house) pay attention to each other 

\begin{figure}[h]
  \centering
  \includegraphics[width=3.3in]{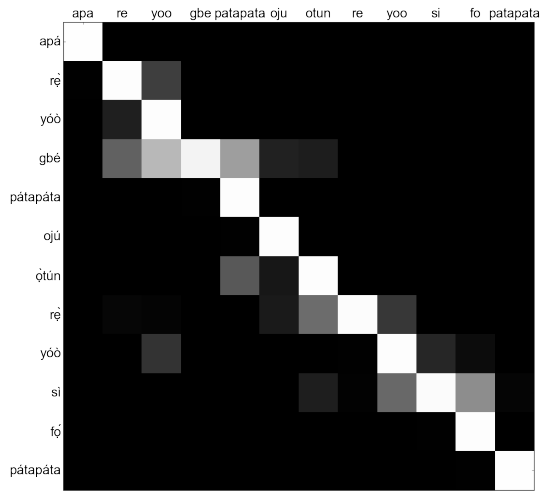}
  \caption{Attention weight matrix (longer)}
  \label{fig:attention_weights_patapata}
\end{figure}

In Figure \ref{fig:attention_weights_patapata}, we provide a longer example, this time noting the greater distances between related words. For example, the second prediction of \textbf{y{\'o}{\`o}} (word 9) learns from a previous occurrence (word 3). Similarly the fourth word predicted, \textbf{gb{\'e}} \emph{(to carry)} heeds both its left and right context across a span of 6 words, though the correct word in this instance is \textbf{gb\d{e}} \emph{(to dry)}. Lastly, the second possessive determiner \textbf{r\d{\`e}} (word 8), unlike the first (word 2) with a single word context, pays attention to the the noun + adjective compound \textbf{oj{\'u} \textbf{\d{\`o}t{\'u}n}} which expresses possession.

\subsection{Error Analysis}
While performing error analyses on the model predictions, we observed inconsistent diacritizations in the training set. This was especially apparent for words with multiple diacritcized characters but a unique diacritized form, for example \textbf{n{\'i}n{\'u}} (inside) or \textbf{or{\'i}\d{s}{\`i}{\'i}r{\'i}\d{s}{\`i}{\'i}} (all kinds of). Errors in the training set lead to incomplete and erratic diacritizations during inference and while a partial restoration may reduce the ambiguity of text for a human reader, it still presents problems for automatic text processing and search applications. Across the full corpus, to detect incorrect variants of both single and multiple diacritized forms, we can first train a word vector model using fastText \cite{bojanowski2017enriching}. Then given a word that was incorrectly predicted in the test set, we can look at it's nearest neighbours and amend the word forms in the training set that are not valid diacritizations. 

Lastly, regarding the composition of the dataset (Table \ref{tab:training_datasets}), numerous sections contained lengthy passages with a literary style or referenced foreign locations and names that were adapted to a characteristic Yor{\`u}b{\'a} form. To make ADR more suitable as a preprocessing step for end-user TTS, ASR applications or any business usage, it'll be necessary to augment the corpus with more general purpose text.

\section{Conclusions}

We have presented a practical study of attention-based sequence-to-sequence learning approaches for diacritic restoration. Avenues for future work include evaluating previous approaches to Yor{\`u}b{\'a} ADR on this new dataset, growing the training corpus and training superior word embeddings.

We foresee many applications for our work. Firstly, it minimizes the amount of manual correction needed to create a high quality text corpus. Secondly, its use by the largest repositories of Yor{\`u}b{\'a} language material on the web will improve search engine performance for both diacritized and undiacritized queries \cite{asubiaro2014effects}. Finally, the integration of automatic diacritizers into text editors, web browser extensions and mobile phone keyboards will greatly simplify input tasks, allowing users to enter text in plain ASCII and have the correct orthography appear on the screen.
\\
\\
The source code and dataset will be available at \url{https://github.com/Niger-Volta-LTI/yoruba-adr}

\section{Acknowledgements}

We thank lexicographer and linguist, K\d{\'{o}}l\'{a} T\'{u}b\d{\`{o}}s\'{u}n for kindly reviewing the training corpora and providing feedback.

\bibliographystyle{IEEEtran}

\bibliography{mybib}

\end{document}